\documentclass[11pt]{article}

\usepackage{amsmath}
\usepackage{amssymb}
\usepackage{mathtools}
\usepackage{authblk}
\usepackage{hyperref}
\usepackage{siunitx}
\usepackage{booktabs}

\usepackage[a4paper, left=3.3cm,top=3.3cm,right=3.6cm,bottom=3.9cm]{geometry}
\usepackage{cmbright}
\usepackage[T1]{fontenc}

\usepackage{enumitem}
\setitemize{label=\scriptsize{$\blacksquare$} }
\setenumerate{label=(\alph*) }

\usepackage{multicol}
\usepackage{multirow}
\usepackage{algpseudocode}

\newcommand*{\edot}{\,\cdot\,}

\newcommand*{\R}{\mathbb{R}}

\newcommand*{\by}{\mathbf{y}}

\newcommand*{\bbf}{\mathbf{f}}
\newcommand*{\bbg}{\mathbf{g}}

\newcommand*{\NN}{\mathcal N}
\newcommand*{\feature}{\mathcal F}
\newcommand*{\class}{\mathcal S}

\newcommand*{\TT}{\mathcal{T}}

\newcommand*{\UU}{\mathbb U}
\newcommand*{\VV}{\mathbb V}
\newcommand*{\WW}{\mathbb W}

\newcommand*{\Xin}{\mathbf{X}}
\newcommand*{\Xout}{\mathbf{Y}}

\renewcommand{\phi}{\varphi}
\renewcommand{\epsilon}{\varepsilon}

\DeclarePairedDelimiter{\norm}{\lVert}{\rVert}

\DeclareMathOperator{\relu}{ReLU}

\allowdisplaybreaks

\title{Image Based  Fashion Product Recommendation\\
with Deep Learning}

\author[1]{Hessel Tuinhof}
\author[2]{Clemens Pirker}
\author[3]{Markus Haltmeier}

\affil[1]{InvoiceFinance; Netherlands\authorcr
E-mail: \texttt{hessel@invoicefinance.nl}\vspace{1em}}

\affil[2]{Department of Strategic Management, Marketing and Tourism\authorcr
University  of Innsbruck, Austria\authorcr
E-mail: \texttt{clemens.pirker@uibk.ac.at}\vspace{1em}}

\affil[3]{Department of Mathematics\authorcr
University of Innsbruck, Austria\authorcr
E-mail: \texttt{markus.haltmeier@uibk.ac.at}}

\date{July 17, 2018}

\begin{document}

\maketitle

\begin{abstract}
We develop a two-stage deep learning framework that recommends fashion images based on other input
images of similar style. For that purpose, a neural network classifier is used as a data-driven,
visually-aware feature extractor. The latter then serves as input for similarity-based recommendations
using a ranking algorithm. Our approach is tested on the publicly available Fashion dataset.
Initialization strategies using transfer learning from larger product databases are presented.
Combined with more traditional content-based recommendation systems, our framework can help to
increase robustness and performance, for example, by better matching a particular customer
style.
 

\end{abstract}

\section{Introduction}\label{sec:intro}

Identifying products a specific customer likes most can significantly increase the earnings of a
company~\cite{schafer2001commerce}. Clearly, recommending suitable products in E-commerce increases
the probability of a customer's purchase. Additionally, offering too many products can reduce the
probability that a potential customer performs a purchase at all. Finally, knowing and subsequently
targeting customer preferences increases the medium- and long-term commitment of the customer to the
company, which is a key factor to profitability~\cite{dick1994customer,srinivasan2002customer}.
Prior studies demonstrate that recommendation engines help consumers to make better decisions,
reduce search efforts and find the most suitable prices~\cite{haubl2006double}.

One possibility to infer knowledge about customer preferences is via specific questioning in
customer surveys. However, this is not always possible and customer responses may not be correct or
sufficient for accurately describing preferences. In this work, we follow a different, data-driven
approach, where customer preferences are automatically extracted from available information on the
customer. More specifically, we focus on fashion products and develop a method that only requires a
single input image to return a ranked list of similar-style recommendations.

\subsection{Proposed recommendation system}\label{ssec:proposal}

The proposed recommendation system operates in a two-stage mode. In the first step, we train a
convolutional neural network (CNN) to solve specific image classification tasks. The trained CNN is
then used as a problem-specific feature extractor, where the features serve as inputs for the
ranking system. While in this paper we work with fashion products, similar recommendation systems
can be employed for other product categories as well.

Image data provides a wealth of information on a visually-aware feature level, e.g.\ edges and color
blobs. Plenty of image processing techniques exist to extract such low-level
features~\cite{prince2012computer}. Deep learning provides a technique to extract hidden
higher-level features by composing several convolutional layers. Therefore CNNs are a natural choice
to provide fashion product recommendations based solely on image data. Compared to classical
content-based recommendation, which is mainly based upon descriptive metadata like manually
annotated product tags or user reviews, our approach relies on visual information.

\subsection{Relation to previous work}\label{ssec:prev}

There are at least two main approaches for product recommendations: collaborative filtering and
content-based filtering. Whereas the former relies on historical user-item interactions, the latter
tries to relate user profiles and item descriptors. A recent deep learning approach is the neural
collaborative filtering framework proposed in~\cite{he2017neural}, which generalizes the matrix
factorization technique used extensively in collaborative filtering methods. Others
like~\cite{he2016vbpr} employ a hybrid approach, where a matrix factorization based predictor is
combined with a deep learning model that extracts visual features as well as latent non-visual user
features. A recent thorough overview on deep learning-based recommender systems can be found
in~\cite{zhang2017deep}.

The success of CNNs for computer vision tasks like object classification, detection and
segmentation~\cite{goodfellow2016deep} gives reason to decouple classical product recommendation
solutions from its extensive user-item interaction data usage requirement. Therefore our method uses
product image data, which, for example in E-commerce, is readily available. This also allows to
mitigate the cold start problem of collaborative filtering and classical content-based recommender
systems. Closely related to our approach are the works~\cite{chen2017image,shankar2017deep}. Due to
the high degree of subjectivity related to fashion articles, general recommender systems usually
perform poorly in fashion recommendation tasks. We show that recommendation systems purely relying
on visual features are reasonable as they are able to provide highly visually appealing
recommendations of similar style.  This can also be helpful in the case of new customers, where no
historical user data is yet available. It can also be integrated in existing content-based systems,
for example, to account for a particular or desired style of a customer, or to address the
cold-start problem.

\subsection{Outline}\label{ssec:outline}

The remainder of the paper is structured as follows. Section~\ref{sec:methods} presents the proposed
product recommendation method. In particular, we give details on the used network architectures, the
used ranking algorithm and describe the Fashion dataset. In Section~\ref{sec:results} we present
some numerical results. The paper concludes with a short discussion in Section~\ref{sec:conclusion}.

\section{Methods}\label{sec:methods}

\subsection{Fashion dataset}\label{ssec:data}

Throughout this paper, we work with a subset of the publicly available
Fashion\footnote{\url{http://imagelab.ing.unimore.it/fashion_dataset.asp}}
dataset~\cite{manfredi2014complete}. In order to obtain high-quality ground-truth labels for
category type and texture attributes, we design a labeling questionnaire on the crowdsourcing
platform CrowdFlower\footnote{\url{www.crowdflower.com}}. Every image is labeled by a total maximum
of five human operators. To be a valid label at least three human operators have to agree. Each
labeling task consists of five images to be labelled, one of which is a simple test image. If a
human operator fails a test more than twice, she is no longer allowed to continue. Separate datasets
for category and texture classification have been created.

\begin{figure}[htb!]
 \centering
 \begin{minipage}{\columnwidth}
 \includegraphics[width=0.49\columnwidth]{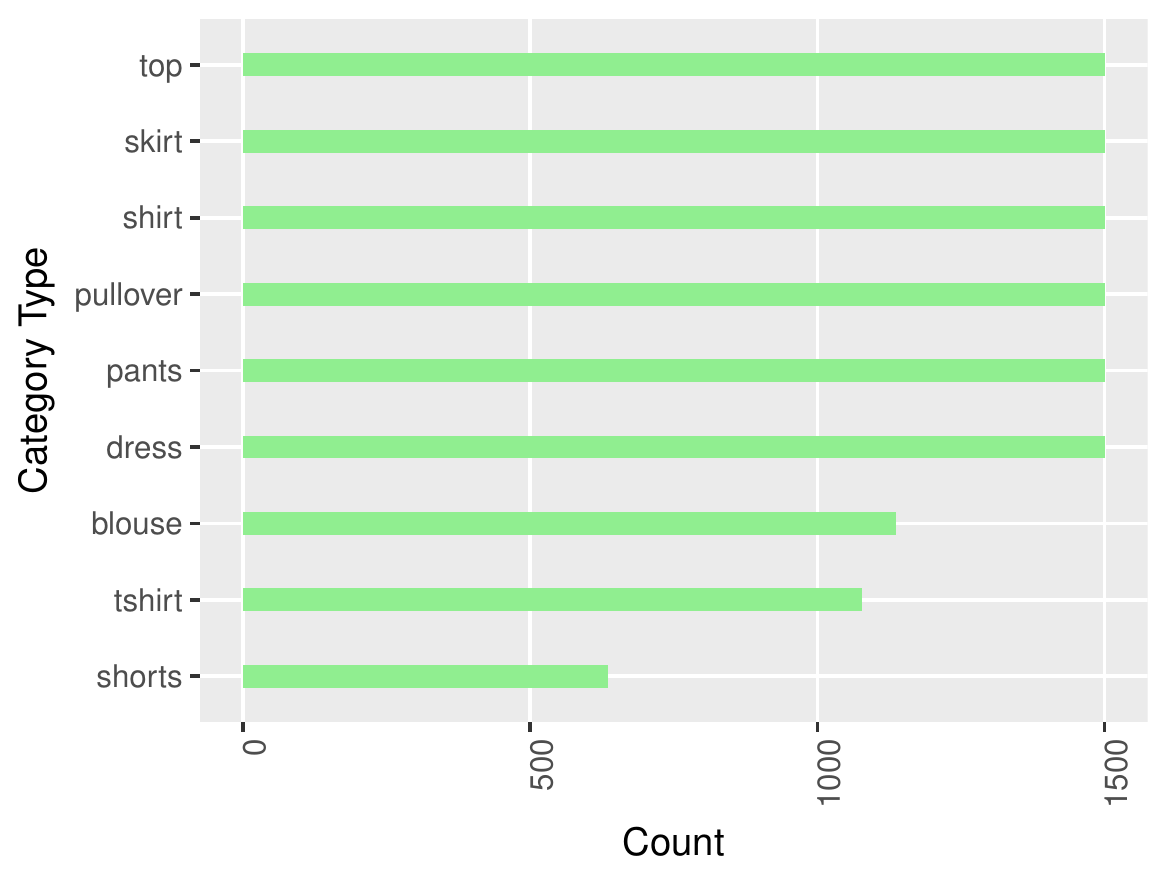}
 \includegraphics[width=0.49\columnwidth]{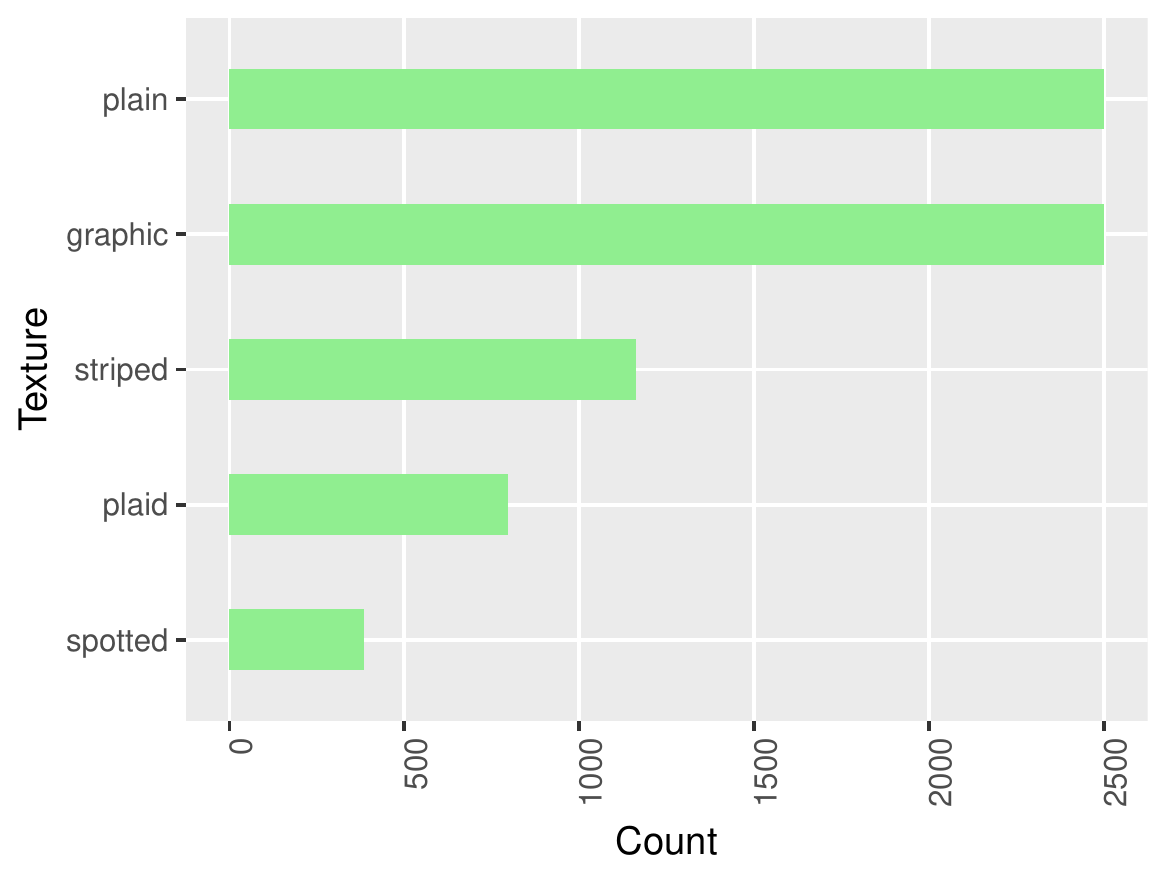}
 \end{minipage}
 \caption{Frequency distributions for the category type and texture attributes
 datasets created from the Fashion dataset.}\label{fig:frequencydistributions}
\end{figure}

The used class labels for category types are blouse, dress, pants, pullover, shirt, shorts, skirt,
top, T-shirt. For the texture attributes we use the labels graphic, plaid, plain, spotted, striped.
Figure~\ref{fig:frequencydistributions} shows the frequency distributions for the two datasets. The
category type dataset contains \SI{11851}{} and the texture attributes dataset \SI{7342}{} images.
Further characteristics can be found in Table~\ref{tab:fashiondata}.

\begin{table}[htb!]
 \centering
 \begin{tabular}{l|ccccc}
 \toprule
 & \multicolumn{2}{c}{Classification} & \multicolumn{3}{c}{Samples} \\
 \cline{2-6}
 Dataset & Type & No. & Total & Train & Val \\
 \midrule
 Fashion Category & Multinomial & 9 & 11,851 & 9,480 & 2,371 \\
 Fashion Texture & Multinomial & 5 & 7,342 & 5,873 & 1,469 \\
 \bottomrule
 \end{tabular}
 \caption{Summary of the datasets created from the Fashion
 dataset. The third column indicates the total amount of class labels for the respective
 dataset.}\label{tab:fashiondata}
\end{table}

\subsection{Proposed framework}\label{ssec:system}

Our method composes of a trained CNN classifier used as image feature extractor and a modification
of the $k$-nearest neighbors ($k$-NN) algorithm used for ranking in feature space.

\begin{itemize}
\item \textbf{Classification via CNNs:}
In the first step, we train separate CNNs to predict the category and texture type. Each of the CNNs
can be written as
 \begin{equation} \label{eq:nn}
	 \NN_i(\WW_i,\edot) \triangleq \class_i(\VV_i, \feature_i(\UU_i,\edot) ) 
	 \;\: \text{ for } i = 1, 2 \,.
\end{equation}
Here $\WW_i = (\UU_i, \VV_i)$ are weight vectors, $\class_i(\VV_i,\edot)$ are fully connected
softmax output layers that actually perform classification and $\feature_i(\UU_i,\edot)$ are the
CNNs without the last layer. The latter are used as feature extractor.

\item \textbf{Ranking in feature space:}
After training and evaluating the performance of these classifiers, we remove the softmax output
layer $\class_i$ of each model. The remaining CNNs are then concatenated and $\feature =
[\feature_1, \feature_2]$ is used to extract the feature vector $\feature(\Xin)$ of any input image
$\Xin\in \R^{N \times N}$. We then use the $k$-NN algorithm to search for the closest items to
$\feature(\Xin)$ in feature space.
\end{itemize}

Details on the employed CNNs and the $k$-NN algorithm for ranking are presented below.

\subsection{Network architectures}\label{ssec:deep}

A wealth of CNN architectures are available today. In this section we briefly discuss the two
architectures that we use in our work: AlexNet and batch-normalized Inception (BN-Inception). The
AlexNet and BN-Inception are both  standard architectures and well established.  AlexNet has been
chosen as a benchmark to compare against deeper, more complex networks like the BN-Inception.
AlexNet consists of 8 layers and BN-Inception of 34. Both use an image of size 224x224 as input.

Two important contributions of the AlexNet~\cite{krizhevsky2012imagenet} are popularizing usage of
the non-saturating rectified linear unit activation function, $\relu(x)\triangleq\max(0,x)$, and introducing
a normalization layer after the $\relu$ activation. Empirical results show that the normalization
layer improves the generalization ability of the network. The BN-Inception~\cite{ioffe2015batch} is
an extension of the GoogLeNet architecture~\cite{szegedy2015going}, which allows deeper and wider
CNNs by mapping the output of a layer to several layers at once. The output of these parallel layers
is then again concatenated. The proposed batch normalization extension addresses the internal
covariate shift problem. The latter describes the problem that the latent input distribution of
every hidden layer constantly changes, because every training iteration updates the weight vector
$\WW_i$. Batch normalization also has a regularization effect.

\subsection{Network training}\label{ssec:train}

In order to adjust $\NN_i(\WW_i, \edot)$ to the particular classification task, the weight vector
$\WW_i$ is selected depending on a set of training data $\TT_i \triangleq \{(\Xin_n,
\Xout_n)\}_{n=1}^{N_i}$. For this purpose, the weights are adjusted in such a way, that the overall
error of $\NN_i(\WW_i, \edot)$ made on the training set is small. This is achieved by minimizing the
error function
\begin{equation} \label{eq:err}
 	E(\WW_i) \triangleq
 	\sum_{n=1}^{N_i} d(\NN_i(\WW_i, \Xin_n),\Xout_n)
	+ \lambda \lVert \WW_i \rVert ^2\,,
\end{equation}
where $d$ is a distance measure that quantifies the error made by the network function $\NN_i(\WW_i,
\edot)$ for classifying the $n$-th training sample.

To stabilize the weight computation in~\eqref{eq:err}, we add a $L^2$-regularization term $\lambda
\lVert \WW_i \rVert ^2$ with regularization parameter $\lambda\geq0$. As is common for
classification with neural networks, we use the cross entropy for the loss function $d$. The actual
minimization of~\eqref{eq:err} is performed by stochastic gradient descent.

\subsection{Ranking by $k$-NN}\label{ssec:knn}

The $k$-NN algorithm can be used as simple ranking algorithm. For that purpose, consider the feature
space $\R^p$ and denote with $d_2(\bbf,\bbg) = \norm{\bbf-\bbg}_2$ the Euclidean distance of two
feature vectors. Let $\{\bbf_1,\ldots,\bbf_m \}$ be a training set of feature vectors. A $k$-NN
algorithm then solves some regression or classification task at $\bbf \in \R^p$ using the $k$
closest training features. This can be implemented by first computing an enumeration
$\pi(\bbf)\colon\{1,2,\ldots,m\}\to\{1,2,\ldots,m\}$ satisfying $d_2(\bbf,\bbf_{\pi(\bbf)(i)}) \leq
d_2(\bbf,\bbf_{\pi(\bbf)(i+1)})$. We use the permutation $\pi(\bbf)$ as ranking output for the input
feature $\bbf$. To reduce memory requirements of the $k$-NN ranking, we use an implementation that
employs a balltree search~\cite{omohundro1991bumptrees}.

\section{Results}\label{sec:results}

In this section we present results for the image classification and similarity recommendation with
the proposed framework.

\subsection{Pretraining}\label{ssec:pre}

To overcome difficulties arising from the relative small size of the Fashion dataset, we use the
concept of transfer learning~\cite{goodfellow2016deep,tajbakhsh2016convolutional}. For that purpose,
we pretrain the classification models on a larger dataset (namely, the DeepFashion Attribute
Prediction\footnote{\url{http://mmlab.ie.cuhk.edu.hk/projects/DeepFashion/AttributePrediction.html}}
dataset,~\cite{liu2016deepfashion}) containing \SI{289222}{} garment images. A full summary of the
dataset can be found in Table~\ref{tab:deepfashiondata}.

\begin{table}[t!]
 \centering
 \begin{tabular}{l|ccccc}
 \toprule
 & \multicolumn{2}{c}{Classification} & \multicolumn{3}{c}{Samples} \\
 \cline{2-6}
 Dataset & Type & No. & Total & Train & Val \\
 \midrule
 DeepFashion Category & Multinomial & 46 & 289,222 & 231,377 & 57,845  \\
 DeepFashion Texture & Multinomial & 156 & 111,405 & 89,124 & 22,281 \\
 \bottomrule
 \end{tabular}
\caption{Summary of the datasets created from the DeepFashion Attribute Prediction dataset used for
 pretraining. The third column indicates the total amount of class labels for the respective
 dataset.}\label{tab:deepfashiondata}
 \end{table}

For pretraining we use AlexNet and BN-Inception architectures. For the AlexNet we
minimize~\eqref{eq:err} with stochastic gradient descent using batch size of 64, regularization
parameter $\lambda=0.0005$, learning rate $0.01$ and momentum $0.9$. For training the BN-Inception
we use the ADAM~\cite{kingma2014adam} algorithm with batch size of 32, $\lambda=0$, and learning
rate $0.001$.  Following~\cite{goodfellow2016deep}, we use early stopping as an efficient
regularization technique to prevent overfitting. We therefore stop training AlexNet/BN-Inception
after 9/8 and 17/13 epochs for the category and texture classification, respectively.

Additional to the cross entropy loss, we use the evaluation metrics accuracy,
\begin{equation}\label{eq:accuracy}
    \operatorname{accuracy}(\by,\hat{\by})\triangleq\frac{1}{N}\sum_{n=1}^N\mathbf{1}_{y_n}(\hat{y}_n)\,,
\end{equation}
and top-$K$ accuracy, which is defined as in Equation~\eqref{eq:accuracy} with a slightly
modified indicator function such that top-$K$ predicted classes are incorporated.
Table~\ref{tab:resultspre-training} shows accuracy, top-$K$-accuracy and loss evaluated on the test
set for both AlexNet and BN-Inception. The BN-Inception achieves higher accuracy and better
generalization ability. Therefore, we only use the BN-Inception architecture for classification on
the Fashion dataset.

\begin{table}[htb!]
 \centering
 \begin{tabular}{l|c|c}
 \toprule
 \multicolumn{1}{c}{Category} & \multicolumn{1}{c}{AlexNet} & BN-Inception \\
 \midrule
 Accuracy & 0.57 & 0.63 \\
 Top-$5$ & 0.79 & 0.84 \\
 Loss & 1.48 & 1.27 \\
 \bottomrule
 \end{tabular}\qquad
 \begin{tabular}{l|c|c}
 \toprule
 \multicolumn{1}{c}{Texture} & \multicolumn{1}{c}{AlexNet} & BN-Inception \\
 \midrule
 Accuracy & 0.28 & 0.32 \\
 Top-$3$ & 0.62 & 0.66 \\
 Loss & 3.00 & 2.82 \\
 \bottomrule
 \end{tabular}
\caption{Pretraining results: The left table depicts results for the category classification
 and the right table for the texture classification.}\label{tab:resultspre-training}
 \end{table}


\subsection{Classification}\label{ssec:classification}

For the final classification models we train the BN-Inception by minimizing~\eqref{eq:err} on the
Fashion dataset with ADAM, where the weights are initialized using the ones from the pretraining
stage. Due to the small size of the Fashion dataset, we add $L^2$-regularization with
$\lambda=0.0001$ to the loss function and also reduce the batch size to 16.

Several image augmentation techniques are applied in order to effectively increase dataset size.
These include random rotations with a maximum rotation angle of $\pm3$ for the category type model
and $\pm8$ for the texture attributes model, random changes of HSL color channels within a range of
$[-6,6]$, a shear transformation with random shear factor within $[-0.25,0.25]$, random aspect ratio
changes within a range of $[0.875,1.125]$ and random vertical flips. The random augmentations are
applied to the training set every epoch anew. This allows to train longer without overfitting too
fast. Following early stopping regularization, we stop training the category type and texture
attributes classification models after 15 and 4 epochs respectively.
Table~\ref{tab:resultsfinetuning} summarizes the final training results. The top-$K$ accuracy metric
is however excluded due the the smaller number of class labels in the Fashion datasets.

\begin{table}[htb!]
 \centering
 \begin{tabular}{l|c|c}
 \toprule
 \multicolumn{1}{c}{} & \multicolumn{1}{c}{Category} & \multicolumn{1}{c}{Texture} \\
 \midrule
 Accuracy & 0.87 & 0.80 \\
 Loss & 0.42 & 0.61 \\
 \bottomrule
 \end{tabular}
\caption{Final BN-Inception classification results on the Fashion datasets for category and
 texture.}\label{tab:resultsfinetuning}
 \end{table}

\begin{figure}[htb!]
 \centering
 \includegraphics[width=0.98\columnwidth]{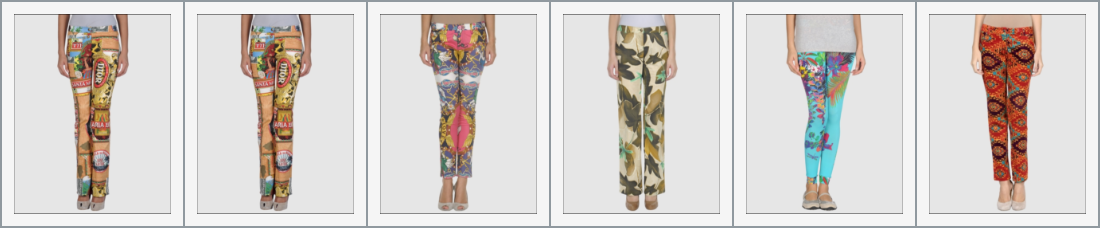}\hfill
 \includegraphics[width=0.98\columnwidth]{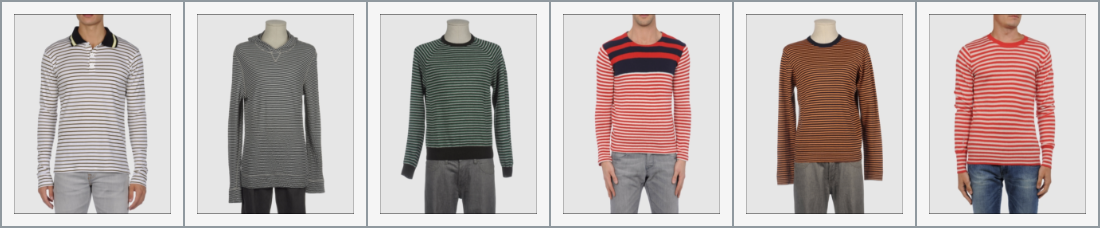}\hfill
 \includegraphics[width=0.98\columnwidth]{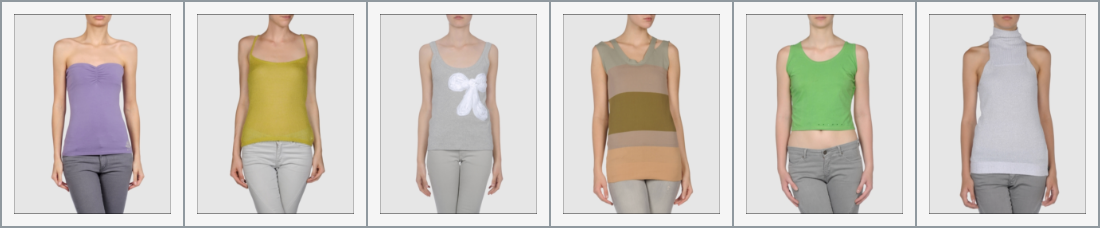}\hfill
 \includegraphics[width=0.98\columnwidth]{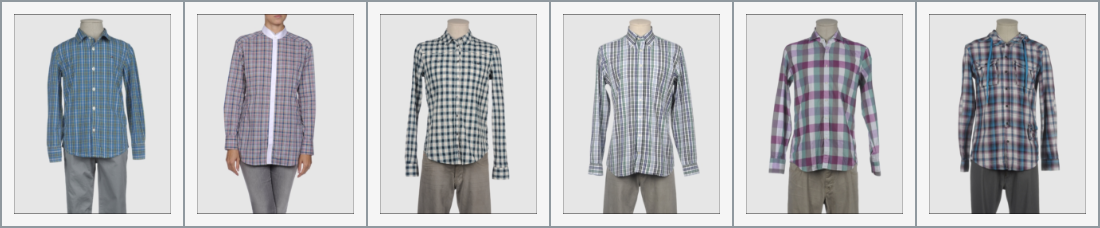}\hfill
 \caption{$k$-NN recommendation ranking. First column displays the query images and
 columns 2--6 display the predicted five nearest neighbors, where column 2 is the most
 similar.}\label{fig:nearestneighbours}
\end{figure}

\subsection{Similarity recommendation}\label{ssec:similarity}

The CNN classifiers are used as feature extractors and return feature vectors $\feature_i(\Xin)$ of
size $d=\SI{1024}{}$ for any input image. The feature extractors are applied to a set of
$n=\SI{19422}{}$ test images. These corresponding feature vectors are concatenated and stacked to
obtain a $n \times 2d$ feature matrix. The $k$-NN ranking algorithm is applied to the feature
matrix. For the recommendation task, it is now sufficient to extract the features from an input
image, submit them to the $k$-NN ranking algorithm and return the top-$k$ matching style
recommendations. In Figure~\ref{fig:nearestneighbours} we present several query images and
corresponding top-$5$ recommendations. Subjectively, the top-5 recommendations indeed look quite
similar to the query images. In the top row a query image from the dataset itself is used. This
corresponding top-5 recommendations demonstrate that if the image appears in the dataset it is
actually most similar to itself. Similar results have been obtained in other performed tests. Other
than that, an implicit objective metric for recommendation quality can be found by means of the
classification accuracies reported in Table~\ref{tab:resultsfinetuning}. The definition of a precise
objective evaluation criterion, however, remains difficult due to the inherent subjectivity of
recommendation quality.  This also makes comparison with other methods quite challenging. The
computationally most time-consuming part in the application of the proposed recommendation system is
the evaluation of the CNN classifiers.

In our implementation, we have implemented the CNNs in MXNet~\cite{chen2015mxnet} using its Python
API. Running on a desktop PC with an Intel i7-6850K CPU and a NVIDIA 1080Ti GPU, the whole image
processing pipeline applied to a given input image only takes fractions of a second. Note that the
potentially time-consuming network training is done before a new input image is provided to the
recommendation system, which therefore allows fast online product recommendation.

\section{Conclusion}\label{sec:conclusion}

We presented a visually-aware, data-driven and rather simple but still effective recommendation
system for fashion product images. The proposed two-stage approach uses a CNN classifier to extract
features that are used as input for similarity recommendations. It can be used, for example, in
E-commerce by allowing customers to upload a specific fashion image and then offering similar items
based on texture and category type features of the customer's uploaded image. Additional feature
extractors, e.g. trained on gender or color classification tasks, can be easily added.  Furthermore,
generalization to other domains makes sense, e.g. music recommendation based on raw music data, but
needs further investigation. Several interesting extensions of our approach are possible.  First, it
would be promising to integrate the two separate training stages into a single one and provide
end-to-end deep learning-based fashion product recommendations. In particular, consideration should
be given to Siamese networks. Additionally, hybrid approaches combining image-based and
content-based systems will be implemented. Finally, it is important to evaluate the customer impact
of our image-based approach and its extensions against other recommender systems through customer
surveys.


\end{document}